\documentclass[letterpaper, 10 pt, conference]{ieeeconf}
\IEEEoverridecommandlockouts
\usepackage{amsmath,amssymb,amsfonts}
\usepackage{algorithmic}
\usepackage{graphicx}
\usepackage{textcomp}
\usepackage{xcolor}
\usepackage{booktabs}
\def\BibTeX{{\rm B\kern-.05em{\sc i\kern-.025em b}\kern-.08em
    T\kern-.1667em\lower.7ex\hbox{E}\kern-.125emX}}

\usepackage[acronym, toc, nonumberlist]{glossaries}
\makeglossaries
\newacronym{LoRA}{LoRA}{Low Rank Adaptation}
\newacronym{PPO}{PPO}{Proximal Policy Optimization}
\newacronym{RL}{RL}{Reinforcement Learning}
\newacronym{SB3}{SB3}{Stable Baselines3}
\newacronym{TRPO}{TRPO}{Trust Region Policy Optimization}
\newacronym{A2C}{A2C}{Advantage Actor-Critic}
\newacronym{CL}{CL}{Continual Learning}
\newacronym{LLM}{LLM}{Large Language Model}
\newacronym{FLOP}{FLOP}{Floating-Point Operations}
\newacronym{SAC}{SAC}{Soft Actor-Critic}
\newacronym{PEFT}{PEFT}{Parameter-Efficient Fine-Tuning}
\usepackage{biblatex}
\bibliography{g12-ref}
\usepackage{placeins}
\usepackage{afterpage}
\let\labelindent\relax
\usepackage{enumitem}
\usepackage{subcaption}

\title{
Memory-Efficient Policy Libraries with Low-Rank Adaptation in Reinforcement Learning
}

\author{\authorblockN{Lyngset, S. V.\textsuperscript{*},
Raanaas, T. V.\textsuperscript{*}, Sveipe, G.\textsuperscript{*},
Nilsen, E. M.\textsuperscript{*}, Torresen, J, Ellefsen, K.O. and Lømo T.\textsuperscript{†}} \\
\authorblockA{Department of Informatics,
University of Oslo}
\thanks{Email: \textsuperscript{†}tobiaslo@ifi.uio.no}
\thanks{\textsuperscript{*}Equal contribution}}

\begin{document}

\maketitle
\thispagestyle{empty}
\pagestyle{empty}

\begin{abstract}
When fine-tuning \glspl{LLM}, there has been success in minimizing both memory usage and computation with \gls{PEFT}, like \gls{LoRA}. In this article, we have explored whether this approach is transferable to the world of robotics and \gls{RL}, allowing learning with reduced memory usage and improved computational performance. Specifically, we focused on a version of multi-task robotics, where a library of specialist policies are created. In such a library memory efficiency is especially important. We used a \gls{PPO} algorithm and fine-tuned a baseline model to different tasks using \gls{LoRA}. Our results demonstrate that, depending on the hyperparameters, \gls{LoRA} can minimize memory usage by a factor of 20-160 compared to full fine-tuning of all layers. This implies a 90-95\% storage saving when deploying a library of many (10-50) specialized policies, which can be the differentiating factor between being able to store the entire library in memory or having to use swap-memory in an applied robotics setting. At the same time, our results indicate that there is no significant difference in the success-rate between full fine-tuning and \gls{LoRA} fine-tuning for the selected tasks.

\end{abstract}

\glsresetall

\section{Introduction}
\begin{figure}[t]
    \centering
    \includegraphics[width=\linewidth]{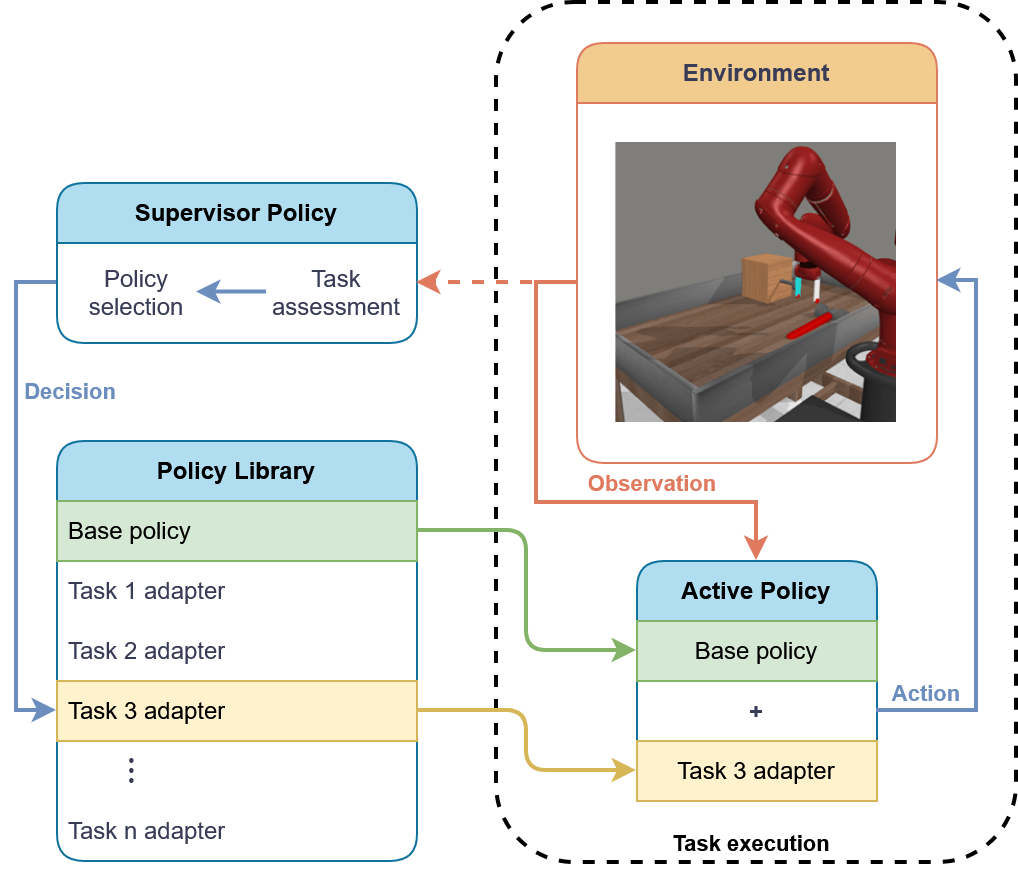}
    \caption{The use case that inspired our approach: A supervisor policy selects from a library of pre-trained specialist policies. This paper focuses on memory efficient training of the specialist policies.}
    \label{fig:model_library}
\end{figure}

Multi-task \gls{RL} aims to find a policy that can solve multiple tasks. This is especially important within robotics, where one robot often needs to solve multiple distinct problems. Multi-task \gls{RL} faces the challenge of catastrophic forgetting~\cite{mccloskey1989catastrophic}, where a policy network forgets the task it has previously been trained for when it is further trained for a new task. Due to catastrophic forgetting, one approach is to train a new policy for each task and save it to a library with specialist policies. Although simple, this approach is storage and memory intensive, as it has to store a separate policy for each task. This paper focuses on memory efficient creation of such a policy-library based on a base policy. Figure \ref{fig:model_library} illustrates how this can be implemented with a supervisor model that selects which specialized model to utilize, similarly to previous work in multi-task \gls{RL}~\cite{xing2024multitaskreinforcementlearningquadrotors}.

\gls{LoRA} \cite{hu2021loralowrankadaptationlarge}, has proved successful in limiting memory usage during fine-tuning, especially in \glspl{LLM}. However, the technique is still relatively unexplored within robotics. \gls{LoRA} is a \gls{PEFT} technique that decomposes the weight updates of a pre-trained model into the product of two much smaller, low-rank matrices. During training, the original model weights are kept frozen and only these low-rank matrices are optimized, effectively constraining the update space to a low-dimensional subspace~\cite{hu2021loralowrankadaptationlarge}. Our experiments show that this technique can also be transferred to the realm of multi-task learning in robotics, enabling  reduced training time, improving knowledge transfer, and limiting memory usage.

\gls{RL} has made great strides in training robots for a single task to a high level of success. Multi-task robotics is not explored as often, but several models with good results have been proposed \cite{lesort2019continuallearningroboticsdefinition}. When using separate policies for each task, a robot must be able to seamlessly switch between several policy networks. For a robot to switch between tasks without a long delay, the memory usage cannot be excessive. If the networks are too large in memory usage, one cannot store them all in memory, and the time it takes to transition into a different task makes the multi-task aspect less viable. 

In this paper, we show the viability of \gls{LoRA} in online \gls{RL}, specifically with \gls{PPO}. In addition, we present a novel use case in robotics using \gls{LoRA} to create a library of specialist models in a more efficient manner. Our experiments show that using \gls{LoRA} instead of fine-tuning all layers greatly reduces the memory needed for every network and might lead to less resource usage during the switching between policies. This is achieved while not decreasing the performance of the policies. We fine-tune every model from a pre-trained base policy trained on a general task. To our knowledge, this has not been shown before. 

\section{Related Works}

\subsection{Expert policy library}

Much work within robotics is focusing on the task of getting a robot to solve multiple tasks and to learn new tasks when needed. This can be split into two sub-fields with overlapping challenges, continual learning and multi-task learning.

For continual learning, also called lifelong learning or Incremental Learning, the focus is on learning new tasks without forgetting what has been learned before. This is called catastrophic forgetting, and is a difficult challenge to solve when working with a single policy network architecture. \cite{wang2024comprehensive, lee2018overcomingcatastrophicforgettingincremental}

In the case of multi-task learning, the focus is more on getting the robot to be able to solve multiple tasks with the same architecture \cite{vithayathil2020survey}. Instead of incrementally adding tasks, all tasks are available at the same time. Using a single policy architecture, this is challenging because different tasks can cause interference with each other, and conflicting gradients during training \cite{yu2020gradient}.

Finding a solution to the problem of continual learning and multi-task learning is an ongoing challenge. For this paper, we instead focus on a simpler solution, using an expert policy library instead of having a single policy architecture \cite{devin2017learning}. Having multiple expert policy networks means that each new task can have its own network, avoiding catastrophic forgetting between tasks. However, with this strategy, the challenge becomes how policies can be trained and stored in an efficient manner. We propose to do this using \gls{PEFT}, as these methods have had great success in large foundational models \cite{hu2021loralowrankadaptationlarge}.

%\subsection{Continual learning}
%\gls{CL} is a machine learning paradigm in which the data distribution and learning objective change throughout time~\cite{wang2024comprehensive}.\gls{CL}, also known as Lifelong Learning or Incremental Learning, has been extensively studied in machine learning as a solution to catastrophic forgetting, the loss of previously acquired knowledge when learning new tasks \cite{lee2018overcomingcatastrophicforgettingincremental}. Early research focused on static datasets and isolated tasks, but recent research in deep learning, representation learning, and \gls{RL} has reignited interest in continual and open-ended learning systems.

%\acrlong{CL} for robotics emphasizes embodied, real-time, and resource-constrained learning, where agents must adapt continuously to changing environments, movement, and sensor experiences. This has led to growing interest in integrating \gls{CL} with developmental robotics, \gls{RL}, and self-supervised exploration to enable autonomous skill acquisition and lifelong adaptation \cite{lesort2019continuallearningroboticsdefinition}.

%Despite rapid progress in the development of \gls{CL} techniques, the problem of adapting to multiple tasks in a robotic setting has not yet been solved. We therefore propose the use of \gls{LoRA} with several networks as a way to train an expanding library of specialist policies for multi-task settings.

\subsection{\gls{PEFT} and RL in Robotics}
Related methods for adapter-based fine-tuning have recently been explored for skill generalization in robotic \gls{RL}. Lu et al.\ investigate the use of adapters to generalize a learned manipulation skill across multiple robotic embodiments \cite{lu2024generalrobot}. In their approach, a disembodied hand is first trained using \gls{SAC} \cite{haarnoja2018soft} to solve a single task, such as opening a drawer. To transfer this abstract skill to robots with differing kinematic constraints, they employ adapter modules, including \gls{LoRA}, updating only adapter layers while freezing the pre-trained policy. They introduce an inverse-kinematics feedback signal that rewards reachable end-effector poses, ensuring the generated trajectories are feasible for the specific robot and task. They report that this method yields improved generalization and success rates compared to fine-tuning of all layers across multiple mobile manipulators, with \gls{LoRA} adapters achieving greater task success in all evaluated robots.

This is related to our work in that both approaches use \gls{LoRA} to specialize a shared baseline policy. Their work focuses on generalizing a single skill from a disembodied hand to multiple robots and a small set of related tasks, such as door-opening and chair-pushing. In contrast, we study \gls{LoRA} for multi-task learning on a single robot and measure how different ranks affect training success, memory usage, and computation when building a library of specialized policies from a shared \gls{PPO} baseline.

Another approach to an adapter based solution is mixture-of-experts (MOE) \cite{mu2025comprehensivesurveymixtureofexpertsalgorithms}. While traditionally used in an LLM environment, it has also been used within the robotics domain. Arora \cite{arora2025multitaskreinforcementlearninglanguageencoded} utilizes smaller policies that represent broken down movements extracted from simple tasks, this is done by using external descriptive information. The gating mechanism then combines the movements into longer and more complex tasks. The idea is that all tasks can easily be sorted and categorized if translated into natural language metadata. The paper also hypothesizes training their gating mechanism by leveraging frozen pre-trained weights.  

Cheng \cite{10111062} uses attention-based-mixture-of-experts which revolves around a backbone network that extracts the common features of each of the selected tasks. This in turn helps the agent understand the environment dynamics. Similar to Arora \cite{arora2025multitaskreinforcementlearninglanguageencoded}, Cheng \cite{10111062} breaks down tasks into simple correlated interactions. For example, identify the object with which you are supposed to interact, using the gripper, and push/pull interactions. After separating all the tasks into specific interactions that correlate with solving different aspects of the said tasks, the attention module grades the expert based on their relevance in solving a given task. 

The difference between their solutions and ours is that they train all tasks in parallel, and their end goal is a single policy that can adapt to various seen and unseen tasks. Their advantage is that the algorithm works on unseen tasks by piecing together a policy based on simpler interactions. A disadvantage is that the success rate is lower than for a solution where you train a policy for each specific task in advance.  Instability and different degrees of success can also be present depending on how the policy is pieced together from the pre-trained tasks.

\subsection{LoRA}

\begin{figure}[t]
    \centering
    \includegraphics[width=.8\linewidth]{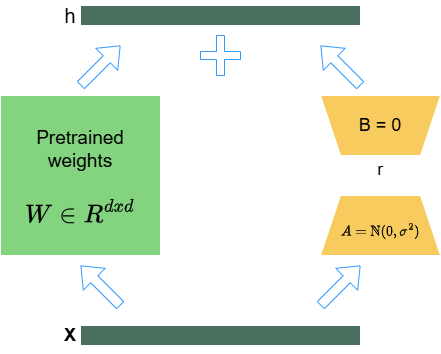}
    \caption{Overview of how LoRA works. The figure is a reconstruction of a figure from the original LoRA paper \cite{hu2021loralowrankadaptationlarge}.}
    \label{fig:lora-fig}
\end{figure}

\acrfull{LoRA} \cite{hu2021loralowrankadaptationlarge} is a technique designed to minimize memory usage during single-task fine-tuning. Unlike full fine-tuning of all layers, which updates all model weights, \gls{LoRA} keeps the pre-trained model parameters frozen (represented as the green square in figure \ref{fig:lora-fig}). Then it injects low-rank trainable matrices (the orange trapezoids in figure \ref{fig:lora-fig}) into the existing layers to handle task specific updates. The \gls{LoRA}-rank defines the inner dimension of the low-rank factorization, where the weight update $\Delta W$ is decomposed into two smaller matrices, A and B. This rank determines the expressive capacity of the adaptation. While a lower rank restricts the update to its most essential features, a higher rank enables the model to capture more complex task-specific information. \gls{LoRA} was created to combat the growing size of \glspl{LLM}. The implementation reduces the number of trainable parameters by roughly a factor of 10,000, improving both memory efficiency and training speed. The approach assumes that parameter updates lie in a low-dimensional subspace of the weight space, which regularizes adaptation and improves stability. This makes \gls{LoRA} highly attractive for \gls{RL}, where training signals are often noisy and data per task limited. Although primarily designed for \glspl{LLM}, the same principle can be applied to robotics. The modular and parameter-efficient updates of \gls{LoRA} could enable robots to adapt previously learned control policies to new tasks with less retraining and storage overhead.

\subsection{PPO}
\gls{PPO} \cite{schulman2017proximalpolicyoptimizationalgorithms} is a policy gradient method for \gls{RL}.
\gls{PPO} provides a surrogate objective for optimizing policy networks by gradient descent in minibatches of trajectories sampled from the environment. This allows efficient optimization while maintaining stability by penalizing large deviations in the policy update. 

\gls{PPO} retains much of the stability of the \gls{TRPO} \cite{pmlr-v37-schulman15} algorithm, but with lower computational cost, approximating the trust-region constraint of \gls{TRPO}, which requires second-order optimization, by a simple first-order clipping mechanism. Empirically, \gls{PPO} has been found to match or exceed \gls{TRPO} in stability and performance in common benchmark tasks \cite{schulman2017proximalpolicyoptimizationalgorithms}.

\gls{PPO} is typically implemented within an actor-critic framework similar to \gls{A2C} \cite{ mnih2016asynchronousmethodsdeepreinforcement, NIPS1999_464d828b}, maintaining a value network (critic) in addition to the policy (actor) to perform advantage estimation, reducing the variance of the policy update. \gls{PPO} has been widely adopted as a baseline algorithm for \gls{RL} \cite{SB3}.

\section{Methods}
\subsection{Environment}
For training and testing our models, we employ the Gymnasium API with the Meta-World environment. Gymnasium provides the framework for the environment we receive from Meta-World. 

Meta-World \cite{metaworld} is an open‑source benchmark designed for multi‑task and meta‑\gls{RL}. It provides 50 different robotic manipulation tasks, such as opening drawers, pushing objects, and pressing buttons. These tasks share a common Sawyer‑arm setup and require combinations of reaching, pushing, and grasping. The tasks include both parametric variations (randomized object and goal positions) and qualitatively different objectives to encourage generalization. The benchmark defines several evaluation regimes: MT1 tests a single policy on many goal variations within one task, while the others (MT10, MT50, ML10 and ML45) require a single policy to solve 10, 45 or all 50 tasks. In the Meta-World paper \cite{metaworld}, state‑of‑the‑art methods achieve modest success on MT10 and struggle a lot on the full MT50, highlighting the difficulty of learning many tasks concurrently \cite{metaworld}.

This paper departs from this intended use: we only employ the MT1 setting, fully training a baseline policy on one task, and then fine‑tuning that policy for other tasks. Meta-World was created to standardize multi-task learning with a single policy for all tasks, while we only train for one single task at a time, focusing on the policy library use-case described in the introduction.

The action space in the Meta-World environments consists of a tuple representing the change in 3D-space of the end-effector and the torque the gripper should apply. The observation space contains the position and orientation of the first object, the second object, and the position of the goal, in addition to all the data from the previous observation. In total, the observation space contains 39 values \cite{metaworld}.

\subsection{Training}
We train a base policy on a single task to be further fine-tuned to create specialist policies for several target tasks, suitable for deployment in a policy library. We selected the \textit{pick-place} task as our base task. \textit{Pick-place} is a task where the robot moves a small cylinder between two random locations. We chose this task for the following reasons:

\begin{enumerate}
    \item It requires 3-axis arm movement, in two stages.
    \item It features randomized primary and secondary goals during training.
    \item It is the task upon which all the reward functions in Meta-World are based \cite{metaworld}.
\end{enumerate}

The variety of target positions ensures that the policy does not learn a static solution, and the variety of movements required are similar to those necessary for many Meta-World tasks. These traits might allow the fine-tuning process to better benefit from knowledge transfer.

The use of the gripper was also found to be a very important skill for the baseline model, as this is something almost all tasks need to do. When we tried to train a baseline-model that had not learned to use the gripper, the \gls{LoRA} fine-tuning struggled. Conceptually, this may be explained by the fact that the fine-tuning process needs to learn to use the gripper from scratch.

We have selected a subset of the Meta-World tasks for testing, shown in figure \ref{fig:results}. The tasks have been selected to represent a variety of the problems available in the Meta-World task library.

Thus, the training pipeline follows the following outline:

\begin{enumerate}
    \item Train the base policy on the base task.
    \item Fine-tune the base policy on a suite of target tasks.
\end{enumerate}

Through trial and error, we found that the hyperparameters listed in table \ref{tab:training} are efficient in our suite of tasks. Unspecified parameters use \gls{SB3} defaults \cite{SB3}. These hyperparameters were used for both the baseline policy and all fine-tuning versions. All fine-tuning policies were trained for a maximum of 2 million steps. However, we used early stopping, which means that training was stopped if performance in 50 evaluation runs was at least 98\%. For evaluation, the model used deterministic action selection. We trained fine-tuned models built on a pre-trained base model for our suite of target tasks using both \gls{LoRA}, with a range of rank values from 1 to 12, and compared with full fine-tuning of all layers.

\begin{table}[t]
    \caption{Training hyperparameters}
    \centering
    \setlength{\tabcolsep}{6pt}
    \begin{tabular}{@{}ll@{}}
    \toprule
    \textbf{Hyperparameter} & \textbf{Value} \\
    \midrule
    Timesteps & 2 000 000 \\
    Rollout buffer & 8192 steps \\
    Learning rate & $10^{-4}$ \\
    Minibatch size & 256 \\
    \gls{PPO} Policy network & 3 dense layers of 400 nodes \\
    \gls{PPO} Value network & 3 dense layers of 400 nodes \\
    \gls{PPO} value function coefficient & 0.4 \\
    \gls{PPO} clip range & 0.15 \\
    \bottomrule
    \end{tabular}
    \label{tab:training}
\end{table}

\begin{figure}[t]
    \centering
    \includegraphics[width=\linewidth]{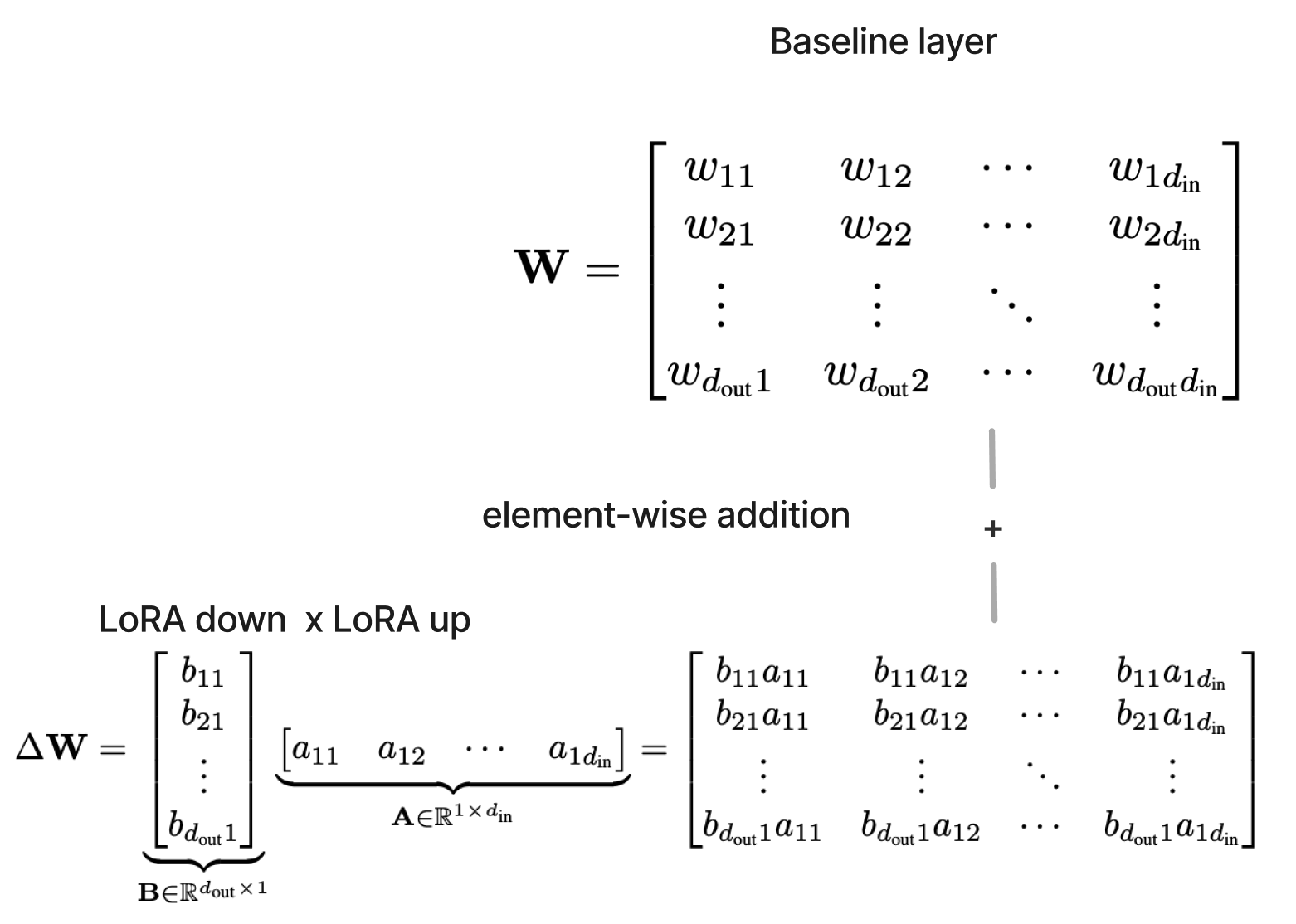}
    \caption{Illustration of a \gls{LoRA} layer with rank=1 showing how a small subset of the parameters can be tuned while still being able to add the weights element-wise.}
    \label{fig:lora_layer}
\end{figure}

\subsection{Architecture}
\begin{figure}[t]
    \centering
    \includegraphics[width=\linewidth]{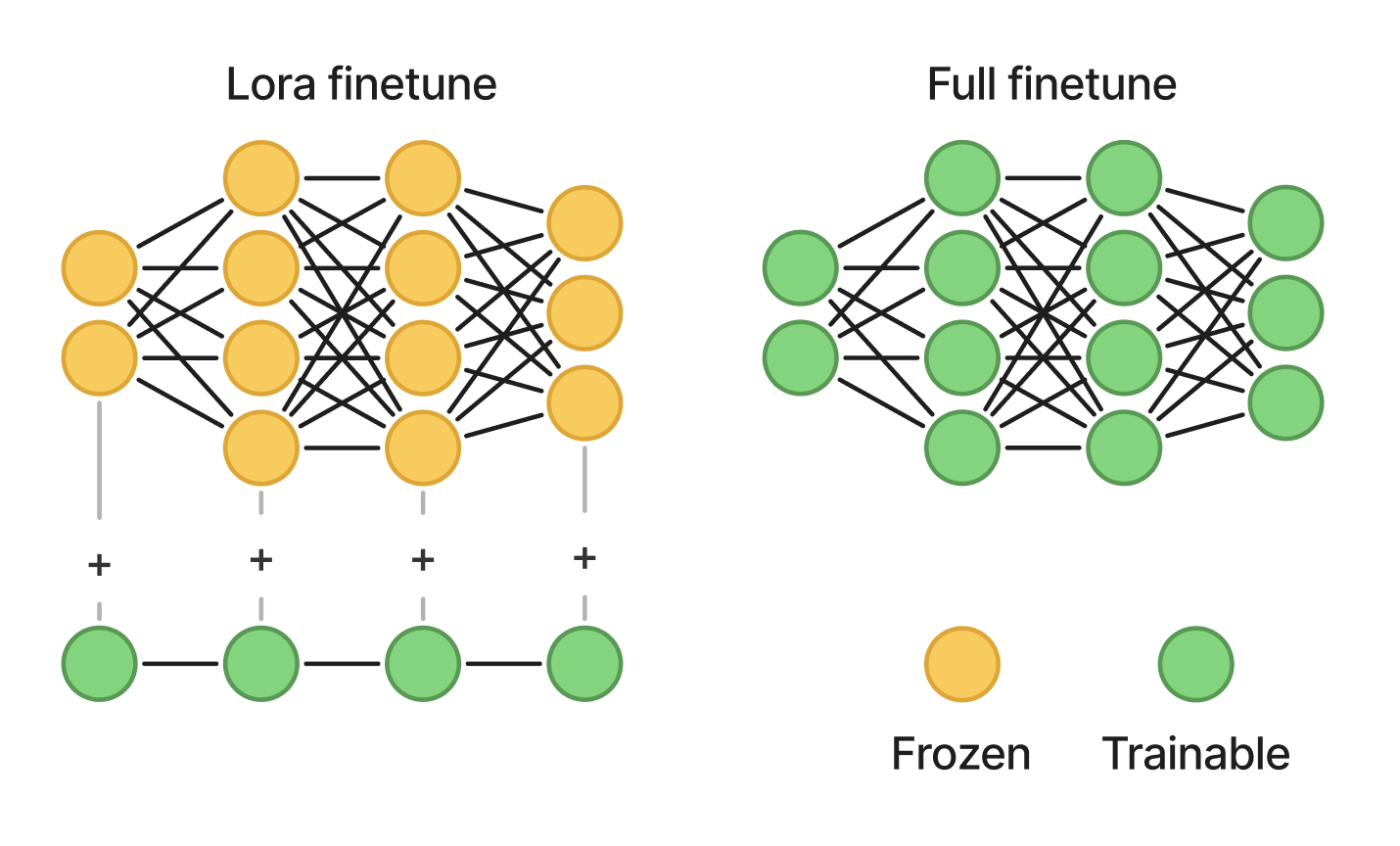}
    \caption{Illustrating how \gls{LoRA} saves memory compared to fine-tuning of all layers by leveraging the base policy and adding its "changes" on top of the base policy weights.}
    \label{fig:architecture}
\end{figure}

We employed the \gls{PPO} algorithm from \gls{SB3} \cite{SB3} to train agents within the Meta-World benchmark suite. To train multiple environments in parallel, we used\gls{SB3}'s VecEnv to use 4 environments simultaneously. The observations and rewards was also normalized. This helps prevent over-fitting and speeds up the training. After a baseline had been trained, either full fine-tuning of all layers or LoRA was used to adopt the policy to the task. Figure \ref{fig:architecture} shows the difference between LoRA and Full fine-tuning. 

For initial training and full fine-tuning setup we use \gls{SB3}´s default actor-critic policy,  while for \gls{LoRA} fine-tuning we have developed our own policy class that inherits from \gls{SB3}'s ActorCriticPolicy class. 

The \gls{LoRA} policy works by wrapping all the layers of a pre-trained model with our LoRaLinear-class that keeps the old frozen weights and creates two new small layers \(LoRA_{down}\) and \(LoRA_{up}\). It overrides the forward function in the following way:
\begin{equation}
\begin{split}
    base_{out} &= base(x)\\
    LoRA_{out} &= LoRA_{up}(LoRA_{down}(x))\\
    h &=  base_{out} + LoRA_{out}
\end{split}
\end{equation}
Figure \ref{fig:lora_layer} illustrates this operation, which allows us to use the \gls{SB3} \gls{PPO} model with very little modification. With our relatively small model, \gls{LoRA} reduces the trainable parameters approximately 20-160 times compared to fine-tuning of all layers. 

\section{Results}

Due to randomness in the different environments, each variation was trained 100 different times. Five different ranks were used, 1, 2, 4, 8, and 12, and were compared with full fine-tuning of all layers. This means that a total of 3600 runs were completed. The following is an explanation of how the metrics for training performance and estimation of computation were calculated. A discussion of the results will be presented in the next section.

\subsection{Training performance}
\begin{figure*}[h]
    \centering
    \begin{subfigure}[b]{0.45\textwidth}
        \centering
        \includegraphics[width=\linewidth]{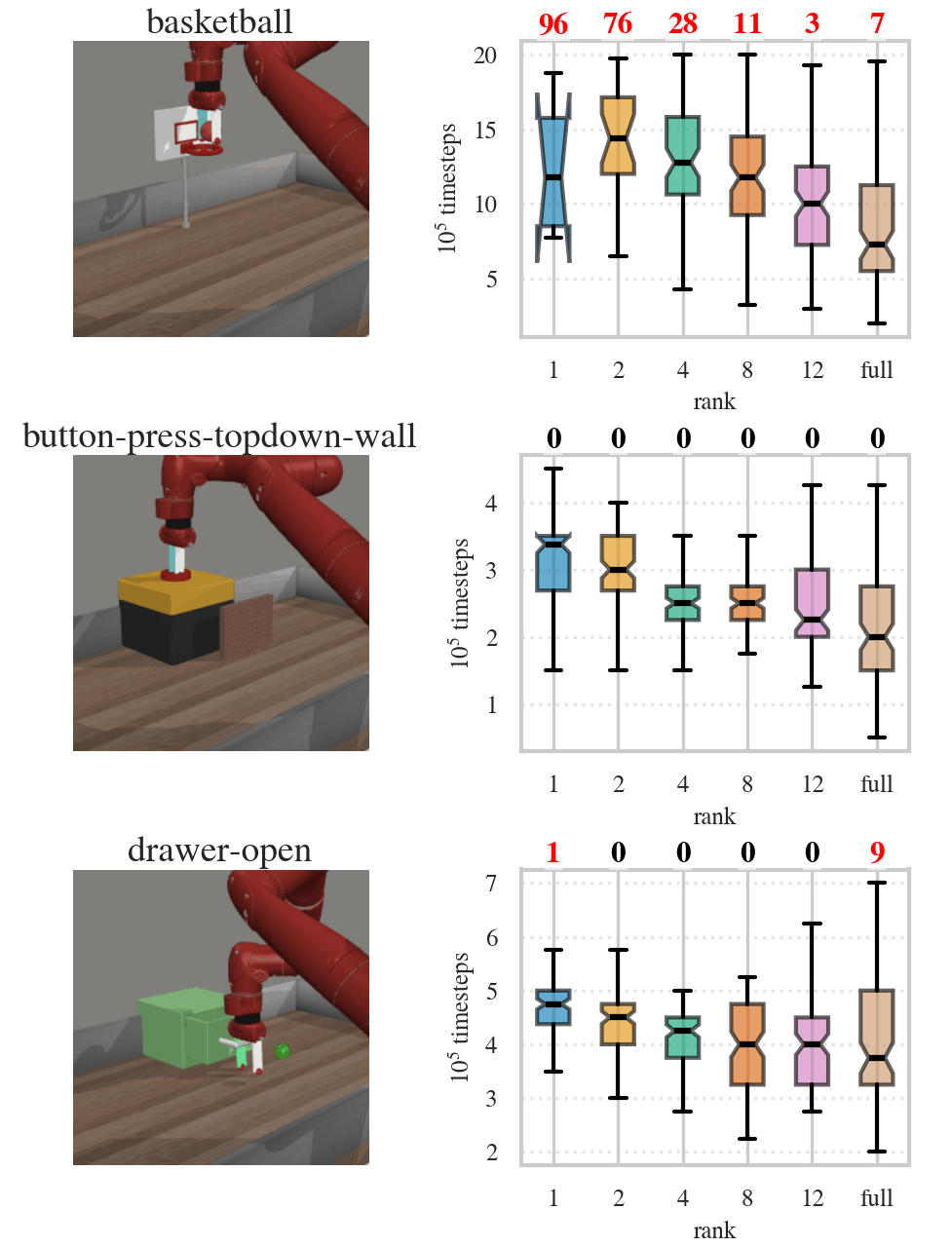}
    \end{subfigure}
    \begin{subfigure}[b]{0.45\textwidth}
        \centering
        \includegraphics[width=\linewidth]{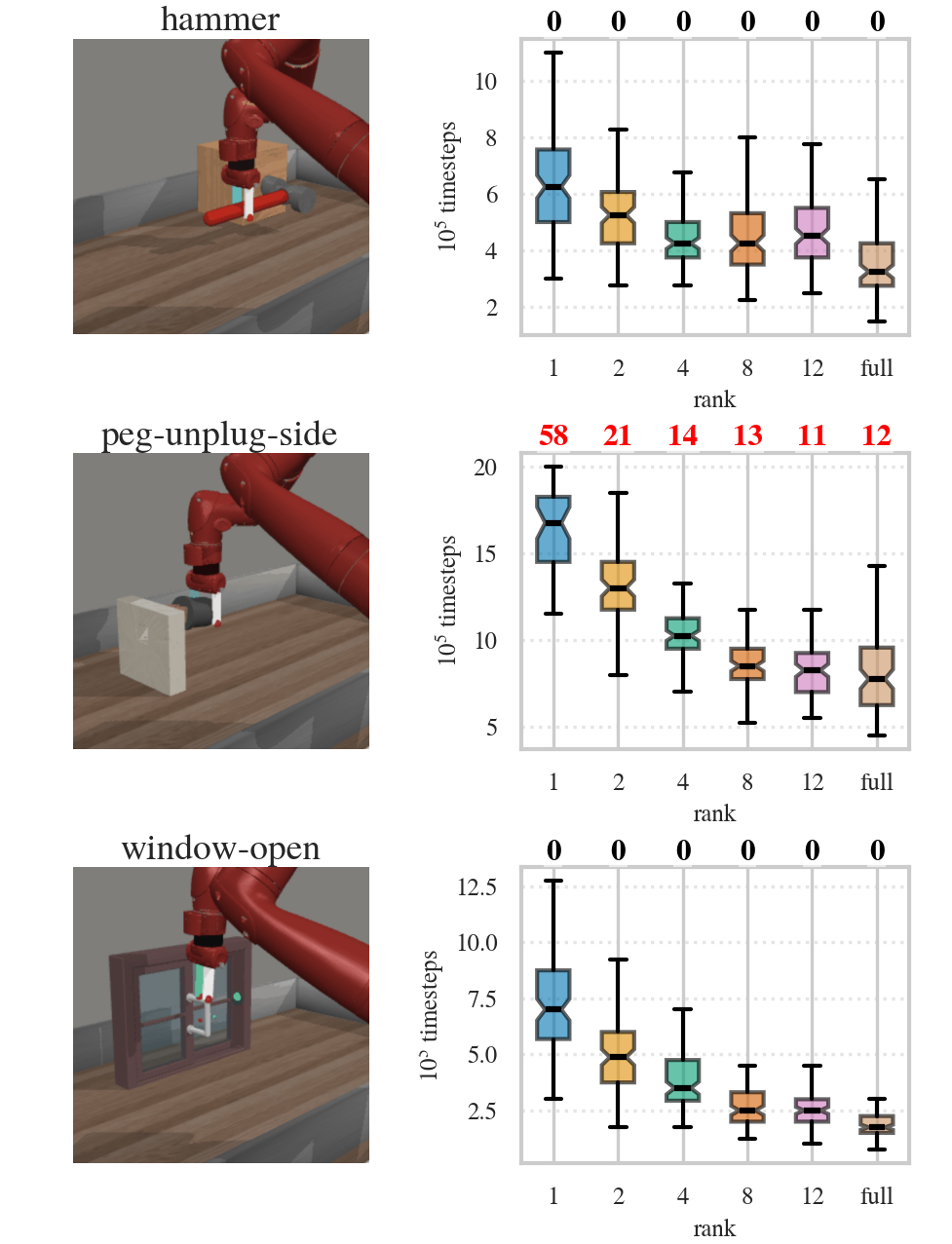}
    \end{subfigure}
    \caption{Boxplots of timesteps until success is achieved (success rate $\geq 0.98$) for 100 runs of the tasks illustrated in the left column. Number of runs that failed to achieve success within the given timesteps ($2 \cdot 10^{6}$) is indicated above the axes. Since it is out of 100 runs, this can also be interpreted as failure rate in percent.}
    \label{fig:results}
\end{figure*}

Achieving early stopping through a 98\% evaluation success rate is defined as successful training in our results, while failing to achieve success in 2 million timesteps counts as a failure. Figure \ref{fig:results} displays the six tasks that were tested, with a boxplot showing the success rate and time steps to success for the different versions of fine-tuning. The boxplot also shows the variability in timestep to success between different runs with the same configuration. Figure \ref{fig:success} provides a more detailed look at the success rate of different rank choices for three selected tasks. Here, the performance is plotted during the training, with the mean performance between runs shown.

\subsection{Computation calculation}
To quantify computational cost, we tracked the number of \glspl{FLOP} per \gls{PPO} update, including both a forward and a backward pass. The total \gls{FLOP} was normalized by the number of environment steps ($n_{\text{steps}} \times n_{\text{envs}}$) to obtain computation per \gls{PPO} environment step:

\[
\text{FLOP}_{\text{step}} = \frac{\text{FLOP}_{\text{update}}}{n_{\text{steps}} \times n_{\text{envs}}}
\]

The total computation used throughout the training (in MFLOP) was obtained by multiplying $\text{FLOP}_{\text{step}}$ by the total number of \gls{PPO} environment steps and dividing by $10^{6}$.
\[
\text{MFLOP}_{\text{total}} = \frac{\text{FLOP}_{\text{step}} \times N_{\text{steps}}}{10^{6}}.
\]
Since all policies share the same architecture, \gls{FLOP} per update depends only on the \gls{LoRA} rank, not on the task.

\section{Discussion}

\subsection{Choice of rank}

\begin{figure*}[t]
    \centering
\includegraphics[width=\linewidth]{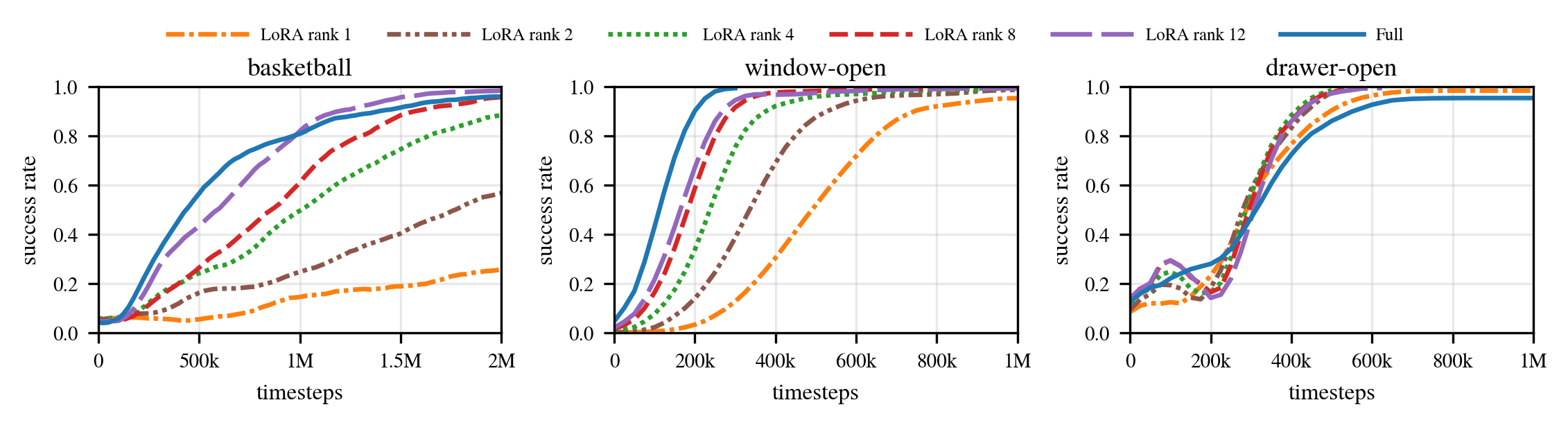}
    \caption{Evaluation success rate achieved by rank, for a selection of tasks.}
    \label{fig:success}
\end{figure*}

The \gls{LoRA} rank parameter determines the complexity of the fine-tuning layers, and also the amount of memory and computation savings achieved. Thus, it should be chosen as low as possible while maintaining the desired level of performance. 

Figure \ref{fig:results} displays the training timesteps required to achieve success for a given rank for our suite of tasks. Observe that for a number of tasks, namely \textit{button-press-topdown-wall}, \textit{hammer}, \textit{drawer-open}, and \textit{window-open}, even the lowest possible rank of 1 consistently achieves success, albeit typically with slightly increased training time. The \textit{basketball} and \textit{peg-unplug-side} tasks might be considered harder tasks. We see that these require a higher rank of about 2 to 8. In particular, rank 1 struggles to solve \textit{basketball}, yielding a high, but not absolute, training failure rate.

Figure \ref{fig:success} gives a more detailed look at how rank impacts the evaluation success rate. The hard tasks, \textit{basketball} and \textit{peg-unplug-side}, resemble the \textit{basketball} plot, with low ranks struggling to achieve a high success rate. However, it is not impossible, as seen in figure \ref{fig:results}.

The easier tasks tend to follow a trend similar to the \textit{drawer-open} or \textit{window-open} plots in figure \ref{fig:success}, with all ranks achieving training success with small differences in training time required. The \textit{window-open} task shows an especially clear distribution of training performance for each rank.

The \textit{drawer-open} task is unusual among our suite of tasks, as it is the only one where the \gls{LoRA} policies achieve higher performance (lower training failure in particular) than the full fine-tune baseline. This can be observed in both figures \ref{fig:success} and \ref{fig:results}. In testing, we observed that this task seems to particularly benefit from knowledge transfer, as \gls{PPO} struggles to solve the task when trained from scratch, but fine-tuned policies achieve a strong success rate. But why it also seems to benefit from more constrained lower rank fine-tuning policies would require further inquiry.

%\begin{figure}[th!]
%    \centering
%    \includegraphics[width=0.44\textwidth]{img/result_grid_half_page.png}
%    \caption{Boxplots of timesteps until success is achieved (success rate $\geq 0.98$) for 100 runs of the tasks illustrated in the left column. Number of runs that failed to achieve success within the given timesteps ($2 \cdot 10^{6}$) is indicated above the axes. Since it is out of 100 runs, this can also be interpreted as failure rate in percent.}
%    \label{fig:results}
%\end{figure}

\subsection{Memory efficiency}
\begin{figure}[t]
    \centering
    \includegraphics[width=0.9\linewidth]{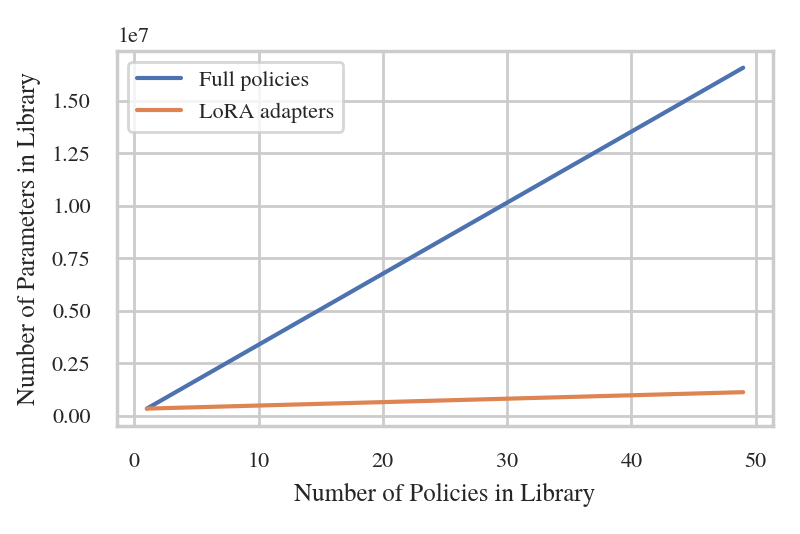}
    \caption{Stored parameter growth of the specialist policy library using \gls{LoRA} rank 8.}
    \label{fig:model_size}
\end{figure}

Full fine-tuning of all layers updates 675,609 parameters at each training step, while \gls{LoRA} updates many fewer. As shown in Table \ref{tab:trainable-params}, \gls{LoRA} reduces the number of trainable parameters by more than 95\% for rank 8 and nearly 99\% for rank 2, resulting in a reduction of roughly 20 to 160 times  the number of trainable parameters compared to full fine-tuning. With a shared base policy of 2.7 MB, storing \gls{LoRA} parameters only yields substantial savings at scale: for ten fine-tuned tasks, the storage drops by about 92\% relative to maintaining separate fully fine-tuned tasks (see Table  \ref{tab:storage}). Each set of \gls{LoRA} parameters adds only 135 KB (rank 8), allowing hundreds of specialized task policies to be efficiently stored and deployed on limited robotic hardware. The effect of this reduction can be seen in Figure \ref{fig:model_size}.

\begin{table}[t]
    \centering
    \caption{Number of trainable parameters for each setup. The \gls{LoRA} setup is compared to the full fine-tuning.}
    \label{tab:trainable-params}
    \setlength{\tabcolsep}{6pt}
    \begin{tabular}{@{}lcc@{}}
        \toprule
        \textbf{Configuration} & \textbf{Trainable Parameters} & \textbf{Savings (\%)} \\ 
        \midrule
        \gls{LoRA} Rank 1 & 4,082 & 99.4\\
        \gls{LoRA} Rank 2 & 8,160 & 98.8\\
        \gls{LoRA} Rank 4 & 16,316 & 97.6\\
        \gls{LoRA} Rank 8 & 32,628 & 95.2\\
        \gls{LoRA} Rank 12 & 48,940 & 92.8\\
        Full fine-tuning & 675,609 & 0\\ 
        \bottomrule
    \end{tabular}
\end{table}

\begin{table}[t]
  \caption{Storage comparison for model library use case assuming a shared base policy (2.7\,MB) with \gls{LoRA} adapters-only per fine-tune with \gls{LoRA}-rank 8.}
  \label{tab:storage}
  \centering
  \setlength{\tabcolsep}{6pt}
  \begin{tabular}{@{}c|cccc@{}}
    \toprule
    & \multicolumn{2}{c}{\textbf{Total Storage (MB)}} 
    & \multicolumn{2}{c}{\textbf{Savings}} \\
    \cmidrule(lr){2-3} \cmidrule(lr){4-5}
    \textbf{Number of policies} 
    & \textbf{Full Fine-tune} 
    & \textbf{Base + \gls{LoRA}} 
    & \textbf{MB} 
    & \textbf{\%} \\
    \midrule
     2  & 5.4   & 2.8 & 2.6   & 48 \\
     5  & 13.5  & 3.4 & 10.1  & 75 \\
    10  & 27.1  & 4.1 & 23.0  & 85 \\
    50  & 135.5 & 9.5 & 126.0 & 93 \\
    \bottomrule
  \end{tabular}
\end{table}

\subsection{Computation}

\begin{figure}[t]
    \centering
    \includegraphics[width=0.9\linewidth]{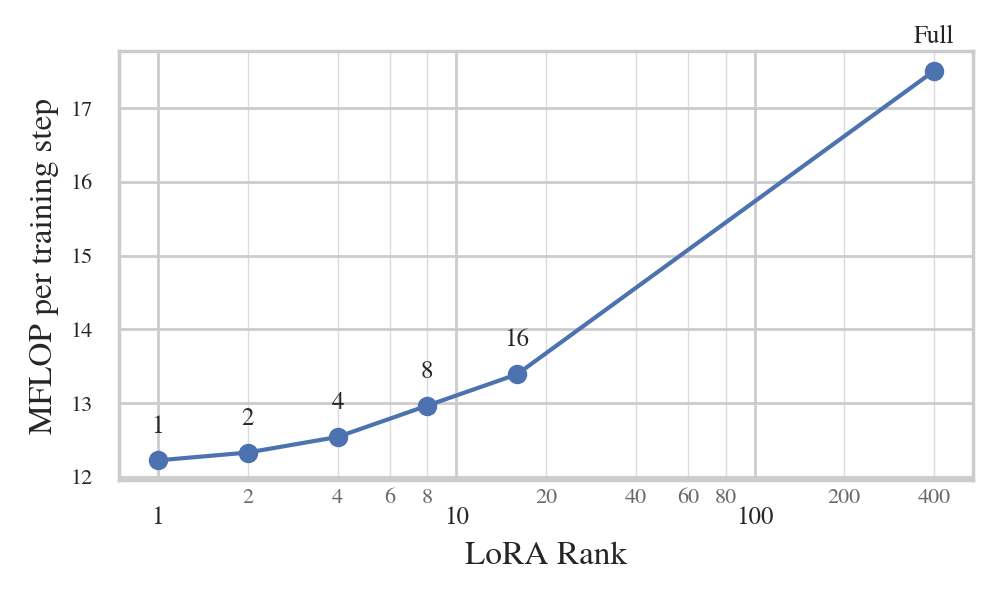}
    \caption{Computation in the backwards step of model training for a given LoRA rank.}
    \label{fig:simple_flops}
\end{figure}

In Figure \ref{fig:simple_flops}, we compare the computational cost of performing a single timestep for different ranks and full fine-tuning of all layers. In general, the results highlight that \gls{LoRA} maintains competitive learning efficiency while substantially reducing computational cost.

In our \gls{FLOP} accounting, we include both forward and backward passes for each \gls{PPO} update. The forward pass is almost identical for \gls{LoRA} and full fine-tuning, with \gls{LoRA} introducing only a very small overhead from the adapter layers, ranging from +0.6\% for rank 1 to +7.3\% for rank 12 in forward \gls{FLOP}.  However, the backward pass is where \gls{LoRA} achieves its computational savings, since only a small number of low-rank parameters are trainable, the backward cost decreases substantially, by approximately 43-50\% compared to fine-tuning all layers in the policy.

Because \gls{PPO} performs multiple forward passes per backward pass during each update, the reduction in backward-pass cost does not translate directly into an equally large reduction in total computation per \gls{PPO} step. This is why the \glspl{FLOP} per \gls{PPO} environment step only decreases by roughly 25–30\% (as shown in Table \ref{tab:flops-per-timestep}), although the backward pass itself is much lighter when using \gls{LoRA}. The observed reduction therefore stems almost entirely from the cheaper backward pass.

It is important to note that the number of \gls{FLOP} reflects the computational cost of training the network and not the entire training process. The most computationally intense part of training is the simulation and data collection. This means that while there are gains in computational performance using \gls{LoRA} for training, the actual impact on speed and resource usage in this simulator is negligible. However, the results may indicate that the results can have a bigger impact in other settings where this imbalance is less present. 

\begin{table}[t]
    \centering
    \caption{\gls{FLOP}$_{step}$ for each fine-tuning configuration.
    \gls{FLOP} are constant across tasks and depend solely on the \gls{LoRA} rank.}
    \label{tab:flops-per-timestep}
    \setlength{\tabcolsep}{6pt}
    \begin{tabular}{@{}lcc@{}}
        \toprule
        \textbf{Fine-tuning Method} & \textbf{\gls{FLOP}$_{step}$} & \textbf{Savings (\%)}\\ 
        \midrule
        \gls{LoRA} Rank 1 & 12,223,888 & 30.2\\
        \gls{LoRA} Rank 2 & 12,329,916 & 29.6\\
        \gls{LoRA} Rank 4 & 12,541,972 & 28.3\\
        \gls{LoRA} Rank 8 & 12,966,084 & 25.9\\
        \gls{LoRA} Rank 12 & 13,390,196 & 23.5\\
        Full fine-tuning & 17,503,460 & 0\\
        \bottomrule
    \end{tabular}
\end{table}

\section{Conclusion}

In this paper, we have explored the use of \gls{LoRA} in online \gls{RL} using \gls{PPO}. This was done in a robotic setting, specifically manipulation tasks. The results show that the performance of fine-tuning using \gls{LoRA} is very similar to full fine-tuning of all layers.

More specifically, \gls{LoRA} fine-tuning gives a memory-usage reduction of 20-160 times depending on the \gls{LoRA} rank (1-12).  Across all tasks tested, \gls{LoRA} closely matched the performance of full fine-tuning while requiring less computation and memory. For simpler manipulation tasks such as \textit{drawer-open}, \textit{hammer}, and \textit{button-press-wall}, all \gls{LoRA} ranks converged rapidly to the same reward plateau as the fully fine-tuned model. More complex tasks such as \textit{basketball} and \textit{peg-unplug-side} suffered in the lower \gls{LoRA} ranks. However, with higher ranks (more complex fine-tuning, less memory saving), it performed at the level with full fine-tuning. This might indicate how much can be inherited from the baseline task.

We propose to use \gls{LoRA} fine-tuning for creating a library of specialist policies for use in multi-task robotics. Using \gls{LoRA}, the memory savings increases with the number of policies saved in the library. Already with 10 specialist policies, the memory saving is 85\% compared with saving the fully fine-tuned policies. This can have a significant effect when switching between several policies in systems, such as mobile robots, with limited resources.

There is a potential for further research on this idea. Our experiments relied on a single baseline policy. testing the effect of different tasks as the baseline would be interesting. In addition more work can be done on a larger set of tasks with greater variety and testing more algorithms like Soft Actor-Critic, Deep Q-Networks or Deep Deterministic Policy Gradient.

\section*{Acknowledgment}
This work is partially supported by the Research Council of Norway as part of the Vulnerability in the Robot Society (VIROS) project (grant no. 288285), the Predictive and Intuitive Robot Companion (PIRC) project (grant no. 312333), Collaboration on Intelligent Machines (COINMAC-2) project (grant no. 309869), through the Centres of Excellence scheme, RITMO (project no. 262762), and through the Norwegian Center for Embodied AI (NCEI) under
grant agreement no. 357451

\printbibliography
\end{document}